\documentclass[sigconf]{acmart-me}
\usepackage{booktabs} % For formal tables
\usepackage{url}
\usepackage{color}
\usepackage{enumitem}
\usepackage{subcaption}
\usepackage{balance}
\usepackage{multirow}

\setcopyright{rightsretained}
\acmDOI{}

\acmISBN{}

\acmConference[MediaEval'21]{Multimedia Evaluation Workshop}{December 12-15 2021}{Online} 
\acmYear{2021}
\copyrightyear{}

\acmPrice{}

\begin{document}
\title{NLP Techniques for Water Quality Analysis in Social Media Content}
%\title{Multi-modal Machine Learning Analysis of Floods}
%\titlenote{Produces the permission block, and
%  copyright information}
%\subtitle{Extended Abstract}
%\subtitlenote{The full version of the author's guide is available as
 % \texttt{acmart.pdf} document}

%%If you want to use multi-column authors that's fine. Comment out the next lines, and then uncomment below.
\author{Muhammad Asif Ayub \textsuperscript{1}, Khubaib Ahmad \textsuperscript{1}, Kashif Ahmad\textsuperscript{2},\\ Nasir Ahmad\textsuperscript{1}, Ala Al-Fuqaha\textsuperscript{2}}
\affiliation{\textsuperscript{1} Department of Computer Systems Engineering, University of Engineering and Technology, Peshawar, Pakistan. \\ \textsuperscript{2} Division of Information and Computing Technology, College of Science and Engineering, Hamad Bin Khalifa University, Qatar Foundation, Doha, Qatar.}
\email{{khubaibtakkar,asifayub836}@gmail.com,{kahmad,aalfuqaha}@hbku.edu.qa, n.ahmad@uetpeshawar.edu.pk}  

%%\authornote{The secretary disavows any knowledge of this author's actions.}
%\affiliation{Institute for Clarity in Documentation, Ohio, USA}
%\email{webmaster@marysville-ohio.com}
%
%\author{Lars Th{\o}rv{\"a}ld}
%%\authornote{This author is the
%%  one who did all the really hard work.}
%\affiliation{The Th{\o}rv{\"a}ld Group, Iceland}
%\email{larst@affiliation.org}
%
%\author{Lawrence P. Leipuner}
%\affiliation{Brookhaven Labs, France}
%\email{lleipuner@researchlabs.org}
%
%\author{Sean Fogarty}
%\affiliation{A Research Institute, Germany}
%\email{fogartys@amesres.org}
%\affiliation{Palmer Research Laboratories, Texas, USA}
%\affiliation{The Th{\o}rv{\"a}ld Group, Iceland}
%\email{jsmith@affiliation.org}

\renewcommand{\shortauthors}{M. Asif et al.}
\renewcommand{\shorttitle}{WaterMM: Water Quality in Social Multimedia}

\begin{abstract}
This paper presents our contributions to the MediaEval 2021 task namely ''WaterMM: Water Quality in Social Multimedia''. The task aims at analyzing social media posts relevant to water quality with particular focus on the aspects like watercolor, smell, taste, and related illnesses. To this aim, a multimodal dataset containing both textual and visual information along with meta-data is provided. Considering the quality and quantity of available content, we mainly focus on textual information by employing three different models individually and jointly in a late-fusion manner. These models include (i) Bidirectional Encoder Representations from Transformers (BERT), (ii) Robustly Optimized BERT Pre-training Approach (XLM-RoBERTa), and a (iii) custom Long short-term memory (LSTM) model obtaining an overall F1-score of 0.794, 0.717, 0.663 on the official test set, respectively. In the fusion scheme, all the models are treated equally and no significant improvement is observed in the performance over the best performing individual model. 
 
\end{abstract}

%\begin{CCSXML}
%<ccs2012>
%  <concept_id>10010520.10010553.10010562</concept_id>
%  <concept_desc>Computer systems organization~Embedded systems</concept_desc>
%  <concept_significance>500</concept_significance>
%  <concept_id>10010520.10010575.10010755</concept_id>
%  <concept_desc>Computer systems organization~Redundancy</concept_desc>
%  <concept_significance>300</concept_significance>
%  <concept_id>10010520.10010553.10010554</concept_id>
%  <concept_desc>Computer systems organization~Robotics</concept_desc>
%  <concept_significance>100</concept_significance>
%  <concept_id>10003033.10003083.10003095</concept_id>
%  <concept_desc>Networks~Network reliability</concept_desc>
%  <concept_significance>100</concept_significance>
%</ccs2012>  
%\end{CCSXML}
%
%\ccsdesc[500]{Computer systems organization~Embedded systems}
%% We no longer use \terms command
%%\terms{Theory}
%
%\keywords{ACM proceedings, \LaTeX, text tagging}

%% Used in some conference proceedings e.g. sigplan and sigchi
%   \caption{This is a teaser}

\maketitle

\section{Introduction}
\label{sec:intro}
In recent years, social media has emerged as a valuable tool and platform to discuss and convey concerns over different challenges and daily life issues \cite{ahmad2019social}. The literature covers a diversified list of societal, environmental, and technological topics, such as racism and hate speech \cite{matamoros2021racism}, public health \cite{naeem2021exploration}, natural disasters and rehabilitation \cite{said2019natural}, and technological conspiracies \cite{hamid2020fake}, discussed in social media outlets. More recently, there have been debates in social networks on environmental issues especially the quality of air and drinking water in different parts of the world. The discussions generally revolve around the topics like strange color, smell, bad taste, and diseases caused by polluted water. This information could help in several ways. For instance, it can serve as valuable feedback for public authorities on the water distribution network. However, extracting information from such informal sources is very challenging. It is possible that social media posts containing water-quality-related keywords do not represent discussions on polluted water. In this regard, Machine Learning (ML) and Natural Language Processing (NLP) techniques could be employed to automatically analyze and filter out irrelevant posts. In order to explore the potential of ML and NLP techniques in this challenging problem, a task namely ''WaterMM: Water Quality in Social Multimedia'' has been introduced in the benchmark MediaEval 2021 competition \cite{andreadis2021watermm}.   

This paper provides a detailed description of the methods proposed by team CSE-Innoverts for the water quality analysis represented in the MediaEval task. The dataset provided for the task covers multi-modal information including textual, visual, and metadata. However, images are available for very few posts. Moreover, the majority of the available images are not relevant. Thus, we mainly focus on textual information by proposing four different solutions as detailed in Section \ref{sec:methodology}.

\section{Proposed Approaches}
\label{sec:methodology}
\begin{figure}[]
%%\label{my1}
\centering
\includegraphics[width=.99\linewidth]{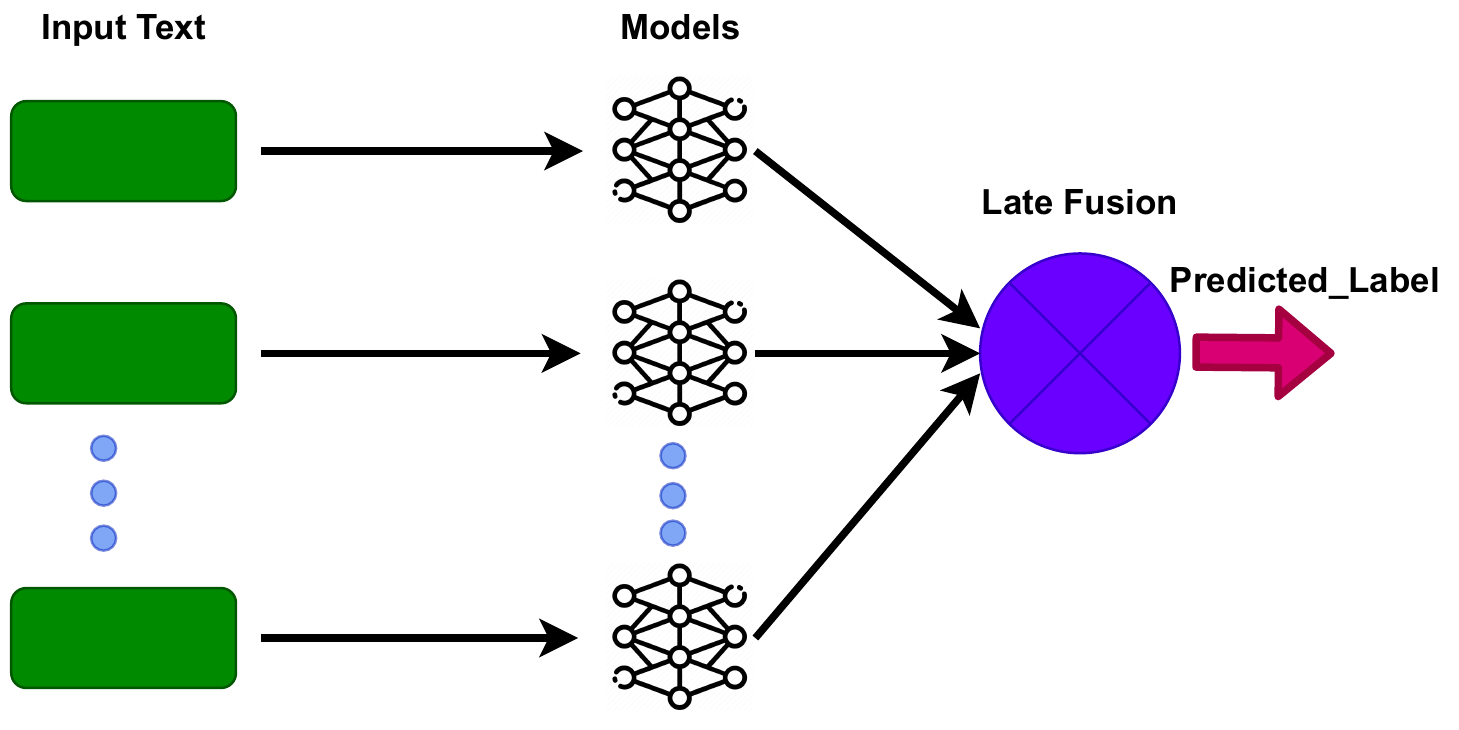}
\caption{Block diagram of the proposed methodology.}
	\label{fig:methodology}
\end{figure}

In total, we submitted 4 different runs by employing three different Neural Networks (NNs) architectures, namely BERT \cite{devlin2018bert}, XLM-RoBERTa \cite{liu2019roberta}, and LSTM, individually and jointly in a late fusion scheme. Run 1 is based on the late fusion where we jointly employed the models by aggregating the classification scores obtained with the individual models. Figure \ref{fig:methodology} provides the block diagram of the proposed methodology for Run 1. Run 2, Run 3, and Run 4 are based on the individual models namely BERT, XLM-RoBERTa, and LSTM, respectively. The details of the individual model based solutions are provided below. 

\begin{itemize}
    \item \textbf{BERT-based Solution (Run 2)}: In this proposed solution, we rely on a pre-trained BERT model, which is fine-tuned on the data development set provided by the task organizers. Before proceeding with fine-tuning the model, necessary pre-processing is performed, using \textit{Tensorflow} libraries, to bring the data in the required form to be used for training the model. Since it is a binary classification task, we used \textit{Binary Cross entropy} loss function with \textit{Adaptive Moments (Adam)} optimizer. 
    \item \textbf{XLM-RoBERTa-based Solution (Run 3)}: In this approach, we rely on the multilingual pre-trained XLM-RoBERTa model that is fine-tuned on the development set. As a first step, the input text is tokenized in the pre-processing phase. A pre-trained model is then fine-tuned on the pre-processed data using Adam optimizer with a binary cross-entropy loss function. 
    \item \textbf{LSTM-based Solution (Run 4)}: In this approach, we rely on a custom LSTM model. The model is composed of three layers including an input, LSTM, and output layer. We used this model as a baseline for our experiments. However, the model obtained encouraging results on the development and was thus utilized in the fusion scheme.
\end{itemize}

%%%%%%%%%%%% Structure Block Diagram %%%%%%%%%%%%%%%%%%%%%%%%%%%%%%%%%%%%%%%
%%%%%%%%%%%%%%%%%%%%%%%%%%%%%%%%%%%%%%%%%%%%%%%%%%%%%%%%%%%%%%%%%%
%%%%%%%%%%%%%%%%%%%%
We also cleaned the data before feeding into the models by removing URLs, account handles, emojis, and unnecessary punctuation. Moreover, in all the proposed solutions, we used an up-sampling technique to balance the dataset. 

\section{Results and Analysis}
\subsection{Evaluation Metric}
For the evaluation of the proposed methods, we used four different metrics, namely (i) accuracy, (ii) micro precision, (iii) micro recall, and (iv) micro F1-score. Precision, recall, and f1-scores are the official metrics while accuracy has been used as an additional metric for the evaluation of the methods on the development set.

\subsection{Experimental Results on the Development Set}
Table \ref{Dev_results} provides the experimental results of our proposed solutions on the development set. To this aim, a separate validation set composed of 1,810 samples is used. Run 1 represents our fusion-based solutions while Run 2, Run 3, and Run 4 represent our solutions based on the individual models namely BERT, RoBERTa, and LSTM, respectively. 
On the development set, overall better results are obtained with the BERT-based solution obtaining an overall F1-score and accuracy of 0.950 and 0.929, respectively. The least performance in terms of F1-score and accuracy are observed for RoBERTa. 
% \usepackage{multirow}
\begin{table}[]
\caption{Evaluation of our proposed solutions on the development set in terms of precision, recall, f1-score, and accuracy.}
\label{Dev_results}
\begin{tabular}{|l|l|l|l|l|}
\hline
\textbf{Runs} & \textbf{Precision} & \textbf{Recall} & \textbf{F1-Score} & \textbf{Accuracy} \\ \hline
Run 1 & 0.950 & 0.925 & 0.938 & 0.914\\ \hline
Run 2& 0.949 & 0.950 & 0.950 & 0.929 \\ \hline
Run 3 & 0.862 & 0.900 & 0.881 & 0.836\\ \hline
Run 4 & 0.885 & 0.947 & 0.915 &0.885 \\ \hline
\end{tabular}
\end{table}
%%%%%%%%%%%%%%%%%%%%%%%%%%%%%%
\subsection{Experimental Results on the Test Set}
Table \ref{Test_results} provides the official results on the test set in terms of precision, recall, and f1-score. Overall better results are obtained for BERT among the individual model-based solutions while the least scores are observed for the LSTM based solution. However, interestingly, no significant improvement in the performance for the fusion-based solution over the best-performing individual models-based solution has been observed. One of the possible reasons could be the low-performing models as all the models are treated equally by simply aggregating the obtained posterior probabilities. This limitation could be addressed by using merit-based fusion where weights are assigned to the contributing models based on the performance of the model. 
%%%%%%%%%%%%%%%%%%%%%%%%%%%%%%
\begin{table}[]
\caption{Evaluation of our proposed solutions on the test set in terms of micro precision, recall, and f1-score.}
\label{Test_results}
\begin{tabular}{|l|l|l|l|}
\hline
\textbf{Runs} & \textbf{Precision} & \textbf{Recall} & \textbf{F1-Score} \\ \hline
 Run 1& 0.732 &	0.866 &	0.794\\ \hline
 Run 2& 0.732 &	0.866	& 0.794 \\ \hline
 Run 3& 0.606 &	0.877 & 0.717 \\ \hline
 Run 4& 0.565 &	0.801 & 0.663 \\ \hline
\end{tabular}
\end{table}
%%%%%%%%%%%%%%%%%%%%%%%%%%%%%%
\section{Conclusions and Future Work}
The quantity and quality of the images associated with the social media posts were not good enough to contribute to the task. Thus, we focused on the textual information only by employing several NNs based solutions. In total, four different solutions including a fusion and three individual models based solutions. In the current implementation, we used a simple fusion mechanism by simply aggregating the posterior probabilities obtained with each individual model. 

In the future, we aim to employ more sophisticated fusion schemes by assigning merit based weights to the contributing models. We also aim to make use of the additional information available in the form of metadata in our future fusion-based solutions. 
%\balance
%\balance
%\begin{acks}
%Add any acknowledgements here.
%\end{acks}
\bibliographystyle{unsrt}
\def\bibfont{\small} % comment this line for a smaller fontsize
\bibliography{sample_me20} 

\end{document}